# Active Semi-Supervised Learning using Submodular Functions

**Andrew Guillory**
Department of Computer Science and Engineering
University of Washington
Seattle, WA 98195
guillory@cs.washington.edu

**Jeff Bilmes**
Department of Electrical Engineering
University of Washington
Seattle, WA 98195
bilmes@ee.washington.edu

## Abstract

We consider active, semi-supervised learning in an offline transductive setting. We show that a previously proposed error bound for active learning on undirected weighted graphs can be generalized by replacing graph cut with an arbitrary symmetric submodular function. Arbitrary non-symmetric submodular functions can be used via symmetrization. Different choices of submodular functions give different versions of the error bound that are appropriate for different kinds of problems. Moreover, the bound is deterministic and holds for adversarially chosen labels. We show exactly minimizing this error bound is NP-complete. However, we also introduce for *any* submodular function an associated active semi-supervised learning method that approximately minimizes the corresponding error bound. We show that the error bound is tight in the sense that there is no other bound of the same form which is better. Our theoretical results are supported by experiments on real data.

## 1 BACKGROUND

Assume we are given an undirected weighted graph in the form of a weight matrix $W$ of size $n$ by $n$. Let $V = [n]$ be the set of all nodes. The nodes of the graph have unknown binary labels given by a vector $y \in \{0, 1\}^n$. Our goal is to predict the values of $y$ from $y_L$, the labels for a small labeled subset $L \subseteq V$. Without making any assumptions about $y$ or $W$, this is impossible. A standard assumption is that

$$\Phi(y) \triangleq \sum_{i<j} W_{i,j} |y_i - y_j| \qquad (1)$$

is small (i.e. the cut given by the labels has small cut value). This intuitively corresponds to the assumption that, on average, nearby points have similar labels. If we predict $y' \in \{0, 1\}^n$ the error of our prediction is $||y' - y||^2$. A reasonable goal for a learning algorithm is to guarantee low error when $\Phi(y)$ is small.

In the standard graph based semi-supervised learning problem, $L$ is either (in practice) given or is (for analysis) selected uniformly at random from the set of all nodes. Much theoretical and practical work has considered this setting (Bengio et al., 2006; Blum and Chawla, 2001; Blum et al., 2004) and learning on graphs has emerged as a popular alternative to learning with feature vectors. Feature vectors can also be transformed into a graph through various methods. Typical bounds for learning from a randomly selected label set show error (roughly) decreases like $O(\Phi(y)/|L|)$ or $O(\sqrt{\Phi(y)/|L|})$ depending on assumptions (Blum et al., 2004). A separate line of work has considered mistake bounds for an online setting in which the labels for the nodes of a graph are predicted sequentially in an adversarially selected order (Herbster and Lever, 2009; Cesa-Bianchi et al., 2009).

We consider active learning methods which get to pick $L$. We specifically study batch (i.e. offline) active semi-supervised learning algorithms which pick $L$, receive $y_L$, then make a prediction $y'$ and suffer loss $||y - y'||^2$. This is in contrast to methods which pick the labeled set $L$ adaptively (Afshani et al., 2007) or make predictions online. It is possible in this setting to give simple but non-trivial deterministic bounds which relate error to $\Phi(y)$ and $L$ and which hold for adversarially chosen labels. These bounds are theoretically interesting since they are distinct from probabilistic bounds commonly used for transductive learning. This setting is also of practical interest since it models real world settings in which exploiting graph structure is important, labels may be adversarial, and it is too expensive to acquire labels one-by-one owing to startup overhead. Compared to sequential active learning, batch active learning potentially reduces the cost of acquiring $L$ through parallelization and economies of scale.

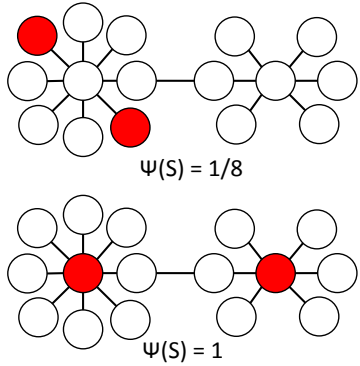

Figure 1: An example of a bad (top) and good (bottom) labeled set according to the $\Psi$ function.

Define the cut function $\Gamma(S)$ for a set of nodes $S$ to be

$$\Gamma(S) \triangleq \sum_{i \in S, j \notin S} W_{i,j} \quad (2)$$

This is the weighted sum of edges crossing $S$ and $V \setminus S$. In previous work (Guillory and Bilmes, 2009), we proposed the following criteria for selecting $L$

$$\Psi(L) \triangleq \min_{T \subseteq (V \setminus L): T \neq \emptyset} \frac{\Gamma(T)}{|T|} \quad (3)$$

This is a variation of the strength of a network (Cunningham, 1985). When $\Psi(L)$ is small, an adversary can cut away a relatively large number of nodes from $L$ without cutting very many edges. When $\Psi(L)$ is large an adversary must cut relatively many edges in order to separate a large number of nodes from $L$. Figure 1 shows an example of this. Intuitively, the minimization over $T$ can be thought of as an adversary selecting which nodes disagree with our predictions. In previous work (Guillory and Bilmes, 2009), we show that $\Psi(L)$ is in fact related to prediction error.

**Theorem 1** (Guillory and Bilmes (2009))**.** *For any labeled graph and $L \subset V$, if we predict*

$$y' = \operatorname*{argmin}_{\hat{y} \in \{0,1\}^n : \hat{y}_L = y_L} \Phi(\hat{y})$$

*then*

$$||y - y'||^2 \leq 2 \frac{\Phi(y)}{\Psi(L)}$$

In other words, if we use the minimum cut method of Blum and Chawla (2001) to predict labels which are consistent with $y_L$ and minimize $\Phi(y')$, then error will be small if $\Phi(y)$ is small. Note that this error bound is deterministic (i.e. holds with probability 1) and in fact holds even when an adversary chooses $y_{V \setminus L}$ after seeing our predictions $y'$. It is possible to get a nontrivial bound in this setting because we only bound error as it relates to the graph smoothness measure $\Phi(y)$ and the graph structure itself–if the adversary chooses labels which disagree with $y'$ and therefore do not match the graph structure then $\Phi(y)$ will be large. This error bound suggests that an active learning method should choose the labeled set to maximize $\Psi(S)$.

A limitation of Theorem 1 is that it is tied to a particular notion of label complexity, graph cut. Theorem 1 can therefore only be applied to *graph based* semi-supervised learning and in particular learning on graphs when $\Phi(y)$ is small. We show that Theorem 1 can be extended by replacing graph cut with an arbitrary symmetric submodular function. This lets us bound error in more general semi-supervised learning problems with other notions of label complexity such as hypergraph cut and mutual information.

Our previous work shows that $\Psi(S)$ can be computed in polynomial time (Guillory and Bilmes, 2009) but leaves open the problem of maximizing $\Psi(S)$. Cesa-Bianchi et al. (2010b) give an algorithm for active learning on tree graphs which is optimal up to constant factors. They show that for tree graphs this method also maximizes $\Psi(S)$ to within a constant factor. When used on general graphs via spanning trees, this method gives good empirical performance (Cesa-Bianchi et al., 2010a) but does not have any (known) theoretical guarantees. We give a method for approximately optimizing $\Psi(S)$ for general graphs. Moreover, our algorithm is for both the graph cut based setting of previous work and our proposed more general setting. We also show the problem is NP-hard and that our bound is tight in the sense there is no better bound of the same form.

## 2 GENERALIZED ERROR BOUND

We begin by generalizing Theorem 1 by replacing the cut function $\Gamma$ with an arbitrary symmetric, submodular objective. Recall that a function $F$ is submodular iff for all $A \subseteq B \subseteq (V \setminus s)$ we have

$$F(A \cup \{s\}) - F(A) \geq F(B \cup \{s\}) - F(B). \quad (4)$$

In other words, $F$ exhibits diminishing returns. An equivalent definition is the following: for all $A, B \subseteq V$

$$F(A \cup B) + F(A \cap B) \leq F(A) + F(B).$$

$F$ is called modular if this inequality holds with equality, monotone if for all $A \subseteq B \subseteq V$ $F(A) \leq F(B)$, and normalized if $F(\emptyset) = 0$. $F$ is called symmetric if $F(S) = F(V \setminus S)$ for all $S$. The cut function $\Gamma$ is submodular, normalized, and symmetric.

The cut function is a natural label complexity function for graph learning applications for which the cut value of the true labels is small. For other applications, other symmetric submodular functions may be better suited. For example, say we are predicting labels for the nodes of a hypergraph (a graph in which edges can connect more than 2 nodes). In this case it is natural to bound error relative to the weight of cut hypergraph edges. This is a symmetric and submodular function. Another example, say that the ground set $V$ is a set of random variables. Symmetric mutual information $\Gamma(S) \triangleq I(S; V \setminus S)$ is a submodular function. With this $\Gamma$, the assumption that the true cut value is small is replaced by the assumption that positively labeled random variables provide little information about negatively labeled random variables. An arbitrary submodular function $F$ can be made into a symmetric, normalized function by taking $\Gamma(S) \triangleq F(S) + F(V \setminus S) - F(V)$. With this construction $F$ can be any submodular function, for example, measuring the rank or coverage of a set $S$.

We overload notation from previous work in order to make the connection to previous results clear. Assume that $\Gamma(S)$ is any symmetric, submodular, normalized, non-negative function. Define $\Psi(S)$ exactly as before (3) in terms of this function. Define

$$\Phi(y) \triangleq \Gamma(V_{y=1}) = \Gamma(V_{y=0})$$

where $V_{y=1}$ is defined to be the subset of $V$ corresponding to items labeled 1 by $y$. $V_{y=0} = V \setminus V_{y=1}$ is similarly defined to be the set of items labeled 0 by $y$. This $\Phi$ is the same as (1) when $\Gamma$ is the cut function and is the number of hypergraph edges cut by $y$ when $\Gamma$ is the hypergraph cut function.

We now state our general error bound for arbitrary symmetric submodular $\Gamma$. Different choices of $\Gamma$ correspond to different assumptions about the true labels $y$ (i.e. different priors, biases), resulting in different bounds which may be useful for different applications.

**Theorem 2.** *Let $\Gamma$ be any symmetric, submodular, non-negative function. For any $y'$ consistent with $y_L$,*

$$||y - y'||^2 \leq \frac{1}{\Psi(L)}(\Phi(y) + \Phi(y'))$$

*Proof.* Let $I$ be the subset of $V$ for which $y'$ is incorrect. Since $y'$ is consistent with $y_L$, $I \cap L = \emptyset$. Thus

$$|I| = \frac{|I|}{\Gamma(I)}\Gamma(I) \leq \frac{1}{\Psi(L)}\Gamma(I)$$

We now argue that $\Gamma(I) \leq \Gamma(V_{y=1}) + \Gamma(V_{y'=1})$, concluding the proof. First note that

$$I = (V_{y=1} \cap V_{y'=0}) \cup (V_{y=0} \cap V_{y'=1})$$

In other words, $I$ is the union of points labeled positive by $y$ but negative by $y'$ and points labeled negative by $y$ but positive by $y'$. Using the submodularity, non-negativity, symmetry and basic set theory we get

$$\begin{aligned}\Gamma(I) &\leq \Gamma(V_{y=1} \cap V_{y'=0}) + \Gamma(V_{y=0} \cap V_{y'=1})\\ &= \Gamma(V_{y=1} \cap V_{y'=0}) + \Gamma(V \setminus (V_{y=1} \cup V_{y'=0}))\\ &= \Gamma(V_{y=1} \cap V_{y'=0}) + \Gamma(V_{y=1} \cup V_{y'=0})\\ &\leq \Gamma(V_{y=1}) + \Gamma(V_{y'=0})\\ &= \Gamma(V_{y=1}) + \Gamma(V_{y'=1})\end{aligned}$$

$\square$

We also get this immediate corollary.

**Corollary 1.** *Let $\Gamma$ be any symmetric, submodular, non-negative function. If $y'$ is chosen to be*

$$y' = \operatorname*{argmin}_{\hat{y} \in \{0,1\}^n : \hat{y}_L = y_L} \Phi(\hat{y})$$

*then*

$$||y - y'||^2 \leq 2\frac{\Phi(y)}{\Psi(L)}$$

In this more general setting, the minimization in Corollary 1 can be performed in polynomial time via submodular function minimization (Fujishige, 2005). In particular the prediction problem corresponds to

$$\min_{y' \in \{0,1\}^n : y'_L = y_L} \Phi(y') = \min_{L_{y=1} \subseteq S \subseteq (V \setminus L_{y=0})} \Gamma(S)$$

Here $L_{y=1}$ is the subset of $L$ labeled 1. This generalizes the mincut method of Blum and Chawla (2001). Note that each different $\Gamma$ implies a different semi-supervised learning strategy for predicting $y$ from $y_L$ (by minimizing $\Gamma(S)$) and a different active learning strategy for selecting $L$ (by maximizing $\Psi(S)$, which is defined in terms of $\Gamma$).

## 3 APPROXIMATION RESULTS

We now consider the problem of finding a minimal size $S$ such that $\Psi(S) \geq \lambda$ for some target $\lambda$.

$$S^* = \operatorname*{argmin}_{S : \Psi(S) \geq \lambda} |S| \quad (5)$$

We also consider constrained maximization

$$S^* = \operatorname*{argmax}_{S : |S| \leq k} \Psi(S) \quad (6)$$

These two problems correspond to picking the smallest labeled set achieving a target error bound and picking the labeled set of size $k$ achieving the best error bound.

**Algorithm 1** Compute $\Psi(S)$
―――――――――――――――――――――――――
$T' \leftarrow V \setminus S$
**repeat**
$\quad T \leftarrow T'$
$\quad \lambda \leftarrow \frac{\Gamma(T)}{|T|}$
$\quad T' \leftarrow \underset{\hat{T} \subseteq (V \setminus S)}{\operatorname{argmin}} \left(\Gamma(\hat{T}) - \lambda|\hat{T}|\right)$
**until** $\Gamma(T') - \lambda|T'| = 0$
**return** $\frac{\Gamma(T)}{|T|}$
―――――――――――――――――――――――――

Although $\Psi(S)$ can be computed in polynomial time it is not immediately obvious how to solve (6) or (5). We first summarize how to compute $\Psi(S)$. Breaking the ratio into a sum

$$\min_{T \subseteq (V \setminus S)} \left(\Gamma(T) - \lambda|T|\right) \quad (7)$$

gives an expression which can be computed in polynomial time for any fixed $S$ and $\lambda$ by solving a submodular minimization problem. When $\Gamma$ is the cut function, this is a mincut problem (Cunningham, 1985). For a fixed $S$, there are only $n - 1$ many critical values of $\lambda$ for which the minimizing set $T \subseteq (V \setminus S)$ changes (Fujishige, 2005). We also have the following result taken from Fujishige (2005)

**Theorem 3** (Fujishige (2005)). *Assume $g(T) \geq 0$, $h(T) \geq 0$, $g(\emptyset) = 0$, $h(\emptyset) = 0$, and $h(T) \neq 0$ for some $T$. $\lambda^* = \min_{T:h(T) \neq 0} \frac{g(T)}{h(T)}$ iff*

$$\forall \lambda \leq \lambda^*, \quad \min_T \left(g(T) - \lambda h(T)\right) = 0$$

*and*

$$\forall \lambda > \lambda^*, \quad \min_T \left(g(T) - \lambda h(T)\right) < 0$$

This theorem motivates a method for computing $\Psi(S)$: if we can find the largest critical value $\lambda^*$ which makes

$$\min_{T \subseteq (V \setminus S)} \left(\Gamma(T) - \lambda|T|\right) = 0. \quad (8)$$

and a *non-empty* $T$ such that $\Gamma(T) - \lambda^*|T| = 0$ we are done. Algorithm 1 shows an iterative algorithm which searches for $\lambda^*$ and $T$ (Cunningham, 1985). This method starts with $\lambda$ set to an overestimate of $\Psi(S)$ (for example $\Gamma(V \setminus S)/|V \setminus S|$) and then iteratively computes a decreased upper estimate stopping when (8) holds. Cunningham (1985) prove this converges after $O(n)$ iterations.

To maximize $\Psi(S)$, we consider (7) as a function of $S$

$$F_\lambda(S) \triangleq \min_{T \subseteq (V \setminus S)} \left(\Gamma(T) - \lambda|T|\right)$$

We first observe that sets which satisfy the constraint $\Psi(S) \geq \lambda$ are exactly the sets which satisfy $F_\lambda(S) = 0$. This result is largely implicit in previous work (e.g. Theorem 3), but it is useful to restate it in terms of functions of $S$ as this is crucial to our approach.

**Lemma 1.** *For any $S \subset V$, $F_\lambda(S) = 0$ iff $\Psi(S) \geq \lambda$*

*Proof.* We first show if $F_\lambda(S) = 0$ then $\Psi(S) \geq \lambda$. Assume for the sake of contradiction that $F_\lambda(S) = 0$ and $\Psi(S) < \lambda$. Therefore there is some $T' \subseteq (V \setminus S)$ : $T' \neq \emptyset$ with $\frac{\Gamma(T')}{|T'|} < \lambda$. Rearranging terms we have $\Gamma(T') - \lambda|T'| < 0$ which contradicts $F_\lambda(S) = 0$.

We now show if $\Psi(S) \geq \lambda$ then $F_\lambda(S) = 0$. Assume for the sake of contradiction that $\Psi(S) \geq \lambda$ and $F_\lambda(S) < 0$. Therefore there is some $T' \subseteq (V \setminus S)$ with $T' \neq \emptyset$ and $\Gamma(T') - \lambda|T'| < 0$. Again rearranging terms we get $\frac{\Gamma(T')}{|T'|} < \lambda$ which contradicts $\Psi(S) \geq \lambda$. □

Note that we can evaluate $F_\lambda(S)$ in polynomial time through submodular function minimization. We can also show that for any fixed $\lambda$, $F_\lambda$ is monotone and submodular. This is desirable because there are efficient algorithms for approximately solving problems like (5) and (6) for monotone submodular functions. Specifically, for monotone submodular functions, simple greedy maximization is approximately optimal (Wolsey, 1982). Therefore if we can reduce maximizing $\Psi(S)$ to maximizing a monotone submodular function, we can approximately maximize $\Psi(S)$. $\Psi(S)$ itself is not submodular (Guillory and Bilmes, 2009) hence the need for the surrogate objective, $F_\lambda(S)$.

The fact that $F_\lambda$ is submodular is a special case of the result that the convolution of a submodular function with a modular function is submodular (Narayanan, 1997). See also Proposition 2 of Cunningham (1985).

**Lemma 2.** *For any $\lambda$, $F_\lambda(S)$ is submodular and monotone non-decreasing*

*Proof.* Monotonicity is obvious: adding elements to $S$ decreases the domain over which the minimization occurs, so $F_\lambda$ increases. A function $F$ is submodular iff $F(V \setminus S)$ is submodular, so it suffices to show that

$$F'_\lambda(S) = F_\lambda(V \setminus S) = \min_{T \subseteq S} \left(\Gamma(T) - \lambda|T|\right)$$

is submodular. Narayanan (1997) defines the lower convolution of $G$ and $H$ to be

$$G * H(S) = \min_{T \subseteq S} \left(G(T) + H(S \setminus T)\right)$$

Narayanan (1997) show that if $G$ is submodular and $H$ is modular then $G * H$ is submodular. We see that $F'_\lambda$ is the lower convolution of $G(T) \triangleq \Gamma(T) - \lambda|T|$, which is submodular, with $H(T) \triangleq 0$, which is modular. □

**Algorithm 2** Minimize $|S|$ s.t. $\Psi(S) \geq \lambda$

---
Define $F_\lambda(S) \triangleq \min_{T \subseteq (V \setminus S)} \Gamma(T) - \lambda|T|$
// Greedily maximize $F_\lambda(S)$
$S \leftarrow \emptyset$
**while** $F_\lambda(S) < 0$ **do**
    $s \leftarrow \underset{\hat{s} \in V}{\operatorname{argmax}}\, F_\lambda(S \cup \{\hat{s}\}) - F(S)$
    $S \leftarrow S \cup \{s\}$
**end while**

---

These two lemmas combine to give an approximation algorithm for (5) (see Algorithm 2).

**Theorem 4.** *Assume $\lambda$ is an integer and $\Gamma$ is integer valued. Applying the greedy algorithm to $F_\lambda$ until $F_\lambda(S) \geq 0$ gives a set $S$ with $\Psi(S) \geq \lambda$ and*

$$|S| \leq (1 + \ln \lambda n) \min_{S:\Psi(S) \geq \lambda} |S|$$

*Proof.* From Lemma 1, (5) is equivalent to

$$S^* = \underset{S:F_\lambda(S) \geq 0}{\operatorname{argmin}} |S|$$

From Lemma 2, $F_\lambda$ is submodular so this is a submodular set cover problem. Wolsey (1982) shows that if $F$ is integer, monotone, submodular, and normalized then the greedy algorithm gives a $1 + \ln F(V)$ approximation for submodular set cover. $F_\lambda(S)$ is submodular, monotone, and integer valued. $F_\lambda(S)$ can be normalized by adding $\lambda n$, giving the result. □

For non-integral but rational $\lambda$ we can rescale $\Gamma$ and $\lambda$ in order to make $\lambda$ an integer. For general $\Gamma$ or $\lambda$ the result of Wolsey (1982) would add an extra normalization term inside the log that is equal to the inverse of the smallest non-zero gain of $F_\lambda$.

In the worst case, the algorithm evaluates $F_\lambda(S)$ $n^2$ times, making the run time of the algorithm $n^2$ that of the submodular function minimization method used. Submodular function minimization can be solved in polynomial time, but currently the fastest known methods in theory require $O(n^6)$ time (Iwata and Orlin, 2009). The minimum norm algorithm of Fujishige et al. (2006) is generally regarded as the fastest known algorithm in practice, but its worst case running time is unknown. If $\Gamma$ is the graph cut function we can use mincut in place of submodular function minimization, greatly improving performance. Standard techniques (Minoux, 1978) can also be used to greatly reduce the number of function evaluations needed.

We can also use Theorem 4 to give a bicriterion guarantee for the maximization problem (6).

**Lemma 3.** *Assuming $\Gamma$ is integer valued, there is a pseudo-polynomial time algorithm finding a set $S$ with*

$$\Psi(S) \geq \max_{S:|S|<k} \Psi(S)$$

*and*

$$|S| \leq (1 + 2 \ln n \Psi(S))k$$

*Proof sketch.* The algorithm simply performs binary search over all possible values of $\Psi(S)$ in order to find the largest value of $\lambda$ such that we can satisfy $\Psi(S) \geq \lambda$ without $|S|$ greater than $(1 + 2 \ln n\lambda)k$. Each of these values is a rational number, so we can rescale $\Gamma$ and $\lambda$ by at most $n$ so that $\lambda$ is an integer. This rescaling introduces an extra factor of at most $n$ inside the log, resulting in the extra factor of 2 outside the log. □

## 4 NEGATIVE RESULTS

We have given an approximation algorithm for finding a minimum size set satisfying $\Psi(S) \geq \lambda$ and a bicriteria approximation for maximizing $\Psi(S)$. These results raise the question: can we do better? We give a partial answer to this question: we show that there are no exact algorithms for either of these problems and no PTAS (polynomial time approximation scheme) for (5). A problem has a PTAS if there is a constant factor approximation algorithm for every constant. Our results do not exclude the possibility of a constant factor approximation, however.

**Theorem 5.** *Assume $\Gamma(S)$ is graph cut. Given as input $W$, $k$, and $\lambda$, determining whether there is a set $S$ with $|S| \leq k$ and $\Psi(S) \geq \lambda$ is NP-complete.*

*Proof.* We show that, assuming $W$ is a binary weight, cubic graph, there is a set $S$ with $|S| \leq k$ and $\Psi(S) \geq 3$ iff the graph has a vertex cover of size $k$. Recall that a cubic graph is a graph where every node has degree 3 and a vertex cover is a set $S \subseteq V$ with either $i \in S$ or $j \in S$ for every $W_{i,j} \neq 0$. The result then follows from the fact that determining if a graph has a vertex cover of size $k$ is NP-complete, even when the graph is restricted to cubic graphs (Garey and Johnson, 1979). Note that in a cubic graph a set $S$ is a vertex cover iff

$$\Gamma(V \setminus S) = 3|V \setminus S| \qquad (9)$$

We first show that if $\Psi(S) \geq 3$ then $S$ is a vertex cover. If $\Psi(S) \geq 3$ then we have from Lemma 1 that

$$\min_{T \subseteq (V \setminus S)} \Big(\Gamma(T) - 3|T| = 0\Big)$$

We argue that one such $T$ achieving this minimum is $V \setminus S$ implying (9) as desired. Assume for the sake of contradiction that

$$\Gamma(V \setminus S) - 3|V \setminus S| > \min_{T \subseteq (V \setminus S)} \Big(\Gamma(T) - 3|T|\Big)$$

Let $T$ be one set achieving $\Gamma(T) - 3|T| = 0$. Consider any $s \in (V \setminus S)$ with $s \notin T$. If we add $s$ to $T$ we increase $\Gamma(T)$ by at most 3 but we increase $|T|$ by 1 so

$$\Gamma(T+s) - 3|T+s| \leq \Gamma(T) - 3|T| = 0$$

Applying this recursively we therefore have

$$\Gamma(V \setminus S) - 3|V \setminus S| \leq \Gamma(T) - 3|T|$$

which is a contradiction.

We now show that if $S$ is a vertex cover then $\Psi(S) \geq 3$, completing the proof. Assume for the sake of contradiction that $\Psi(S) < 3$ and therefore by Lemma 1

$$\min_{T \subseteq (V \setminus S)} \Big(\Gamma(T) - 3|T|\Big) > 0$$

By (9)

$$\min_{T \subseteq (V \setminus S)} \Big(\Gamma(T) - 3|T|\Big) \leq \Gamma(V \setminus S) - 3|V \setminus S| = 0$$

which is a contradiction. $\square$

Note we've shown the result for $\Gamma$ equal to the cut function, but clearly the problem with arbitrary $\Gamma$ is only harder. Since the decision problem is NP-complete, both (5) and (6) are also NP-complete, as polynomial time algorithms for these problems would imply polynomial time algorithms for the decision problem. The reduction used in the proof of Theorem 5 is approximation preserving (specifically it leaves the problem size unchanged), and computing minimum vertex cover on a cubic graph is known to be APX-complete (Alimonti and Kann, 1997), so we also get the following corollary.

**Corollary 2.** *There is no PTAS for (5) unless P=NP.*

Theorem 5 shows that our proposed algorithm is approximately optimal in the sense that (unless P=NP) no polynomial time algorithm can exactly minimize the error bound which it approximately minimizes. A different question is whether or not the error bound itself (Theorem 2) is optimal. In previous work (Guillory and Bilmes, 2009) we remark that a different version of Theorem 1 is tight. Here we show formally in our more general setting that no other error bound of the same form can be strictly better.

**Theorem 6.** *Assume $\Gamma(S)$ is symmetric and submodular and $\Gamma(V) = 0$. Assume that for some function $G(S)$ depending only on $\Gamma$ and $S$ we have*

$$||y - y'||^2 \leq \frac{1}{G(L)}(\Phi(y) + \Phi(y'))$$

*for any $L$, $y$, and $y'$ consistent with $y_L$. We must have $G(L) \leq \Psi(L)$ for every $L$ with $\Psi(L) \neq 0$.*

*Proof.* Assume for some such $G$ we have $G(L) > \Psi(L)$ for some $L$ with $\Psi(L) \neq 0$. Set $y'$ to be a vector of all ones. Note $\Phi(y') = 0$ in this case. Let

$$T = \underset{\hat{T} \subseteq (V \setminus L) : \hat{T} \neq \emptyset}{\operatorname{argmin}} \frac{\Gamma(\hat{T})}{|\hat{T}|}$$

Set $y_i = 1$ for $i \in V \setminus T$ and $y_i = -1$ for $i \in T$. By construction we have

$$\Psi(L) = \frac{\Gamma(T)}{|T|} = \frac{\Phi(y) + \Phi(y')}{||y - y'||^2}$$

We then have $G(L) > (\Phi(y) + \Phi(y'))/||y - y'||^2$ which contradicts our assumption $||y - y'||^2 \leq (\Phi(y) + \Phi(y'))/G(L)$ $\square$

$\Psi(L) \neq 0$ holds under mild assumptions. For example, for the cut function $\Gamma$, $\Psi(L) > 0$ when $L$ contains at least one node in each connected component of $W$.

## 5 EXPERIMENTS

### 5.1 COMPARISON TO BASELINES

We tested our method on a set of benchmark data sets from Chapelle et al. (2006) (also used in previous work (Guillory and Bilmes, 2009)) and two citation graph data sets, Citeseer and Cora, from Sen et al. (2008) (Cora was also used by Cesa-Bianchi et al. (2010a)). In these experiments we set $\Gamma$ to be standard graph cut. We tried our method for maximizing $\Psi(L)$ (Lemma 3) in conjunction with minimum cut prediction (Blum and Chawla, 2001) and also the version of label propagation proposed by Bengio et al. (2006) (using $\mu = 10^{-6}$, $\epsilon = 10^{-6}$, and class mass normalization). Minimum cut prediction is more directly motivated by our theory (when $\Gamma$ is graph cut) but label propagation sometimes works better in practice. We note Theorem 2 still holds for label propagation prediction if we set $L$ to be the set of points on which our predictions agree with the observed labels. Also, in our experiments we use binary search to find a set $S$ with $|S| \leq k$ (as opposed to $|S| < \tilde{O}(k)$), and we do not perform binary search over all rational values of $\Psi(S)$ but instead stop when the relative difference between the upper and lower bounds is less than .0001.

We compare against three baselines: minimum cut prediction with a random labeled set, label propagation prediction with a random labeled set, and the METIS active learning heuristic (Guillory and Bilmes, 2009). For predicting with $k$ labeled points, the METIS heuristic partitions the graph into $k$ parts using the METIS graph partitioning program (Karypis and Kumar, 1999), requests a label for a single random point in each of the $k$ parts, and then labels each part according to that label. This method was found to outperform a variety of other heuristic methods on the benchmark data sets. We average over 1000 runs of the baseline methods for each dataset / label count. For our methods we average over 100 different runs of the final selection(after binary search for $\lambda^*$) on different random permutations of the data.

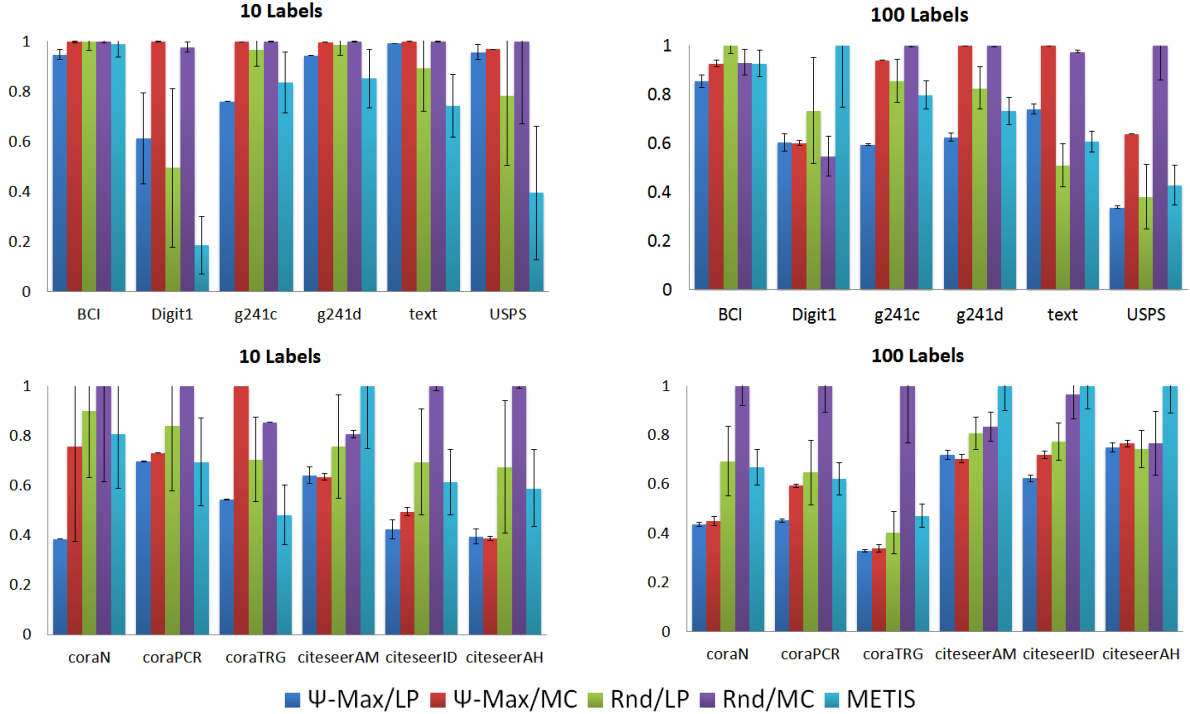

Figure 2: Relative error compared to worst method for different data set / label counts. Top row is $k$-nn graphs, bottom row citation graphs. $\Psi$-max is our method, Rnd is random selection, MC is minimum cut, LP is label propagation, and METIS is a heuristic described in the text. Bars show one standard deviation.

For the benchmark data sets we construct a $k$-nearest neighbor graph with $k = 10$. These data sets are of size $n = 1500$ except for BCI which is of size $n = 400$. We use an unweighted graph for all methods except label propagation for which we used a Gaussian kernel weighted graph ( $W_{i,j} = e^{||x_i-x_j||^2/2\sigma^2}$ ). We set $\sigma$ with a heuristic from Chapelle et al. (2006): we use $1/3$ the average distance to the $k$-th nearest neighbor. We chose to use an unweighted graph for our method as in some preliminary experiments on other data sets we found it can be sensitive to the choice of $\sigma$.

On the citation graph data sets, following the setup used by Cesa-Bianchi et al. (2010a), we use the largest connected component of the graph, group together small classes to form more balanced classes, and perform one-vs-rest binary classification. We do not use the feature vectors for the documents on these data sets–only the citation graph structure. The edges in these data sets are unweighted, and we treat them as undirected. The Cora subset we use is of size $n = 2485$, and the Citeseer subset is of size $n = 2110$. The class groupings we use for Cora are Neural Networks (coraN), Theory / Reinforcement Learning / Genetic Algorithms (coraTRG), and Probabilistic Methods / Case Based / Rule Learning (coraPCR). For Citeseer we use AI / ML (citeseerAM), IR / DB (citeseerID), and Agents / HCI (citeseerAH).

Figure 5.1 shows our results for different data set / label count combinations. On the benchmark data sets, our method combined with label propagation performs better than the others on 6/12 data set / label combinations, but on some problems our method hurts. This is not entirely surprising in light of previous observations made about minimum cut prediction and $k$-nearest neighbors graphs (Blum et al., 2004). We speculate a different choice of graph construction or $\Gamma$ is needed to get more consistent improvement. Graph construction and hyper parameter selection for semi-supervised learning is in general a difficult problem.

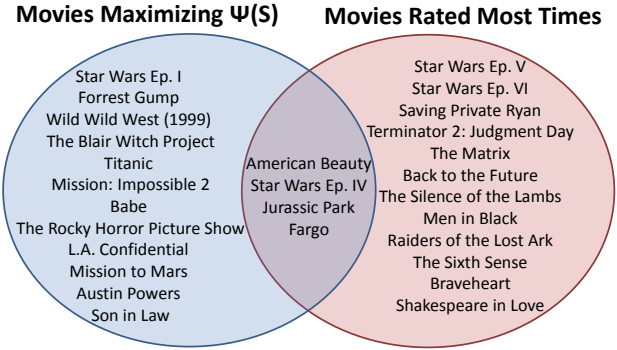

Figure 3: Movies informative about a taste found by our method compared to movies rated most frequently.

However, on the two citation graph data sets where the graph is not constructed by us (i.e. the graph is part of the data), our method has a significant and more consistent benefit. On 9/12 of the classification tasks one of our active learning methods performs best (either the minimum cut or label propagation version), and the difference between our active learning methods and the other methods is often large. On all but one of the classification problems the label propagation version of our method is within 1 percent of the best method. The different variations of Cora (Citeseer) differ only in their labels, so our method selects the same labeled points for each of these variations. This is a feature of our batch setting. Results also confirm our method reduces variance in error. We think our method's strong performance on these natural graph data sets suggests that, when the choice of $\Gamma$ is appropriate for the data, our method performs well.

In a followup experiment we tried running METIS 100 times per trial instead of once, keeping the partition with the lowest cut value. This variation helped some on a few data set / label combinations (the biggest was a 10% relative decrease in error on Digit1 / 10 where METIS already performed well) but overall had very little effect (relative to variance in error) and didn't change trends in the results. We have also begun experimenting with alternative graph constructions and objectives for the benchmark data sets, for example the $\delta$ construction of Blum et al. (2004), but have not yet achieved consistent improvement

## 5.2 CHOOSING MOVIES TO RATE

In this section we discuss a problem setting which uses the added generality of Corollary 1. We consider the problem of training a collaborative filtering system for a new user: which items (e.g. movies) should we ask a new user to evaluate first so that we can then give the user accurate recommendations?

We treat this problem as an active learning problem over a hypergraph by constructing a hypergraph in which nodes are items and edges encode user preferences. We make for each user an edge connecting all movies that the user rated higher than 3 (out of 5) stars and an edge connecting all movies that the user rated lower than 3 stars. With a graph constructed like this, a partition of the hypergraph which cuts few hypergraph edges corresponds to a grouping of movies that is on average consistent with all users' preferences: on average, movies that are liked by the same user will be on the same side of the cut as will movies that are disliked by the same user. More complicated methods could assign different weights to the hypergraph edges or create more than two edges per user.

We set $\Gamma(S)$ to be the hypergraph cut function counting the number of hypergraph edges crossing $S$ and $V \setminus S$. Corollary 1 then suggests a semi-supervised learning method for predicting which items a user likes based on their likes and dislikes for a subset of items $L$. This method guarantees low error so long as the user's true preferences $y$ are mostly consistent with previously seen ratings. This Corollary also suggests that in order to minimize error a new user should rate a set of items $L$ which maximizes $\Psi(L)$.

We tested our $\Psi$ maximization method on the 1 Million Ratings version of the MovieLens data. We use as a ground set all movies which received more than 10 ratings (3233 nodes). The construction method described above then gives 11479 hypergraph edges (about 737000 user/movie connections). We compute hypergraph mincuts using the flow method of Yang and Wong (1996). Figure 3 shows the 16 movies selected by our method in order to minimize $|S|$ subject to $\Psi(S) \geq 2.5$. We compare to the 16 movies which were rated the most number of times.

There is some overlap between the two lists: movies that are rated many times will have high degree in the hypergraph, and therefore may be useful for learning. However, the movies that are selected by our method are more diverse in certain ways and therefore potentially more useful for learning. For example, our method chose movies that are controversial (The Blair Witch Project which was liked/disliked with proportion about 1.18:1 in the data set), a movie considered a cult classic (The Rocky Horror Picture Show), a children's movie (Babe), and even unpopular movies (Son in Law, which received an average rating of 2.67/5). Unpopular movies may be valuable as they can confirm which movies a user does not like. In comparison, the movies rated most often are largely popular movies.

# 6 OPEN PROBLEMS

Our results do not exclude the possibility of a constant factor approximation for either (5) or (6).

The algorithm of Cesa-Bianchi et al. (2010b) for active learning on tree graphs is optimal in a very strong sense: they show that their algorithm makes approximately minimal errors on any tree graph. Our results for more general graphs are not as strong. We only show that there is no strictly better bound of the same form. This is a weaker result since there may be a strictly better bound of a different form.

Our experimental results only consider graph cut and hypergraph cut. We have yet to explore the practical value of other more general symmetric submodular functions. We conjecture that other more general functions can be used to improve performance where standard cut based methods fail (e.g. on some of the $k$-nearest neighbor graphs in our experiments). The use of more general submodular functions will require the use of more general submodular function minimization. Practical submodular function minimization is an active area of research (Stobbe and Krause, 2010; Fujishige et al., 2006).

### Acknowledgments

This material is based upon work supported by the National Science Foundation under grant IIS-0535100, by an Intel research award, and by a Microsoft research award.